\documentclass[10pt,twocolumn,letterpaper]{article}

\usepackage[pagenumbers]{cvpr} 

\def\cvprPaperID{**}
\def\confName{****}
\def\confYear{****}

\usepackage{graphicx}
\usepackage{amsmath}
\usepackage{amssymb}
\usepackage{booktabs}
\usepackage{multirow}

\usepackage{fp}
\usepackage{expl3}[2012-07-08]
\ExplSyntaxOn
\ExplSyntaxOff

\usepackage{amsmath,amsfonts,bm}

\def\x{{x}}

\def\xi{{\x_i}}

\newcommand{\ignorethis}[1]{}

\def\eqref#1{equation~\ref{#1}}

\def\1{\bm{1}}

\def\vc{{\bm{c}}}

\def\vp{{\bm{p}}}

\def\vr{{\bm{r}}}

\def\vt{{\bm{t}}}

\def\vv{{\bm{v}}}

\def\vx{{\bm{x}}}

\def\vz{{\bm{z}}}

\DeclareMathAlphabet{\mathsfit}{\encodingdefault}{\sfdefault}{m}{sl}
\SetMathAlphabet{\mathsfit}{bold}{\encodingdefault}{\sfdefault}{bx}{n}

\newcommand{\ignore}[1]{}

\renewcommand*{\thefootnote}{\fnsymbol{footnote}}

\makeatletter
\DeclareRobustCommand\onedot{\futurelet\@let@token\@onedot}
\def\@onedot{\ifx\@let@token.\else.\null\fi\xspace}

\makeatother

\usepackage[dvipsnames]{xcolor}

\newcommand{\lj}[1]{\textcolor{black}{ #1}}
\newcommand{\wangcanc}[1]{\textcolor{black}{#1}}

\newcommand{\Tref}[1]{Table~\ref{#1}}

\newcommand{\Fref}[1]{Fig.~\ref{#1}}

\newcommand{\sref}[1]{Section~\ref{#1}}

\usepackage[pagebackref,breaklinks,colorlinks]{hyperref}

\usepackage[capitalize]{cleveref}
\crefname{section}{Sec.}{Secs.}
\Crefname{section}{Section}{Sections}
\Crefname{table}{Table}{Tables}
\crefname{table}{Tab.}{Tabs.}
\begin{document}
\title{CLIP-NeRF: Text-and-Image Driven Manipulation of Neural Radiance Fields}
\author{Can Wang \\
City University of Hong Kong \\
{\tt\small cwang355-c@my.cityu.edu.hk}
\and
Menglei Chai\\
Snap Inc.\\
{\tt\small cmlatsim@gmail.com}
\and
Mingming He \\
USC Institute for Creative Technologies\\
{\tt\small hmm.lillian@gmail.com}
\and
Dongdong Chen \\
Microsoft Cloud AI\\
{\tt\small cddlyf@gmail.com}
\and
Jing Liao\footnotemark[1] \\
City University of Hong Kong\\
{\tt\small jingliao@cityu.edu.hk}
}

\maketitle
{
  \renewcommand{\thefootnote}%
    {\fnsymbol{footnote}}
  \footnotetext[1]{Corresponding Author.}
}
\begin{abstract}
We present CLIP-NeRF, a multi-modal 3D object manipulation method for neural radiance fields (NeRF). By leveraging the joint language-image embedding space of the recent Contrastive Language-Image Pre-Training (CLIP) model, we propose a unified framework that allows manipulating NeRF in a user-friendly way, using either a short text prompt or an exemplar image. Specifically, to combine the novel view synthesis capability of NeRF and the controllable manipulation ability of latent representations from generative models, we introduce a disentangled conditional NeRF architecture that allows individual control over both shape and appearance. This is achieved by performing the shape conditioning via applying a learned deformation field to the positional encoding and deferring color conditioning to the volumetric rendering stage. To bridge this disentangled latent representation to the CLIP embedding, we design two code mappers that take a CLIP embedding as input and update the latent codes to reflect the targeted editing. The mappers are trained with a CLIP-based matching loss to ensure the manipulation accuracy. Furthermore, we propose an inverse optimization method that accurately projects an input image to the latent codes for manipulation to enable editing on real images. We evaluate our approach by extensive experiments on a variety of text prompts and exemplar images and also provide an intuitive interface for interactive editing. Our implementation is available at \textcolor{red}{\url{https://cassiepython.github.io/clipnerf/}}
\end{abstract}

\section{Introduction}
\label{sec:intro}

With the explosive growth of 3D assets, the demand for manipulating 3D content to achieve versatile re-creation is rising rapidly. While most existing 3D editing methods operate on explicit 3D representations~\cite{zwicker2002pointshop,ju2007editing,fried2019text}, the recent advances of implicit volumetric representations in capturing and rendering dedicated 3D structures~\cite{park2019deepsdf,mildenhall2020nerf,riegler2020free,jiang2020local,genova2020local,kar2017learning} have motivated the research to benefit the manipulation from such representations. Among these works, neural radiance fields (NeRF)~\cite{mildenhall2020nerf} utilize a volume rendering technique to render neural implicit representations for high-quality novel view synthesis, providing an ideal representation for 3D content. 

Editing NeRF (\textit{e.g.}, deforming the shape or changing the appearance color), however, is an extremely challenging task. First, since NeRF is an implicit function optimized per scene, we cannot directly edit the shape using the intuitively tools for the explicit representations~\cite{schmidt2016state,wang20193dn,wang2020pixel2mesh,uy2020deformation}. Second, unlike image manipulation where the single-view information is enough to guide the editing~\cite{li2020manigan,xia2021tedigan,xu2021text}, the multi-view dependency of NeRF makes the manipulation way more difficult to control without the multi-view information. More recent works propose conditional NeRF~\cite{schwarz2020graf}, which trains NeRF on one category of shapes and enables manipulation via latent space interpolations utilizing the pre-trained models. Based on the conditional NeRF, EditNeRF~\cite{liu2021editing} takes the first step to edit the shape and color of NeRF given user scribbles. However, due to its limited capacity in shape manipulation, only adding or removing local parts of the object is allows. 
In addition to achieving more compelling and complicated manipulation, we seek to edit NeRF in more intuitive ways, such as using a text prompt or a single reference image.

In this paper, we explore how to individually manipulate the shape and the appearance of NeRF based on a text prompt or a reference image in a unified framework. 
Our framework is built on a novel disentangled conditional NeRF architecture, which is controlled by the latent space disentangled into a shape code and an appearance code. The shape code guides the learning of a deformation field to warp the volume to a new geometry, while the appearance code allows controlling the emitted color of volumetric rendering. Based on our disentangled NeRF model, we take advantage of the recently proposed Contrastive Language-Image Pre-training (CLIP) model~\cite{radford2021learning} to learn two code mappers, which map CLIP features to the latent space to manipulate the shape or appearance code. Specifically, given a text prompt or an exemplar image as our condition, we extract the features using the pre-trained CLIP model, feed the features into the code mappers, and yield local displacements in the latent space to edit the shape and appearance codes to reflect the edit. We design the CLIP-based loss to enforce the CLIP space consistency between the input constraint and the output renderings, thus supporting high-resolution NeRF manipulation. Additionally, we propose an optimization-based method for editing a real image by inversely optimizing its shape and appearance codes. 

To sum up, we make the following contributions:
\begin{itemize}
\item We present the first text-and-image-driven manipulation method for NeRF, using a unified framework to provide users with flexible control over 3D content using either a text prompt or an exemplar image. 
\item We design a disentangled conditional NeRF architecture by introducing a shape code to deform the volumetric field and an appearance code to control the emitted colors. 
\item Our feedforward code mappers enable the fast inference for editing different objects in the same category compared to the optimization-based editing method~\cite{liu2021editing}.
\item We propose an inversion method to infer the shape and appearance codes from a real image, allowing editing the shape and appearance of the existing data.
\end{itemize}

\section{Related Work}

\noindent\textbf{NeRF and NeRF Editing.}
The past few years have witnessed tremendous progress in the implicit representation of 3D models with neural networks~\cite{park2019deepsdf,mildenhall2020nerf,riegler2020free,jiang2020local,genova2020local,kar2017learning}. Among them, NeRF~\cite{mildenhall2020nerf} is a representative one, which encodes a continuous volume representation of shape and view-dependent appearance in the weights of an MLP network. NeRF has been gaining more and more popularity because of its strong capability in capturing high-resolution geometry and rendering photo-realistically novel views. The success of NeRF has also inspired many follow-up works that extend the NeRF to dynamic scenes~\cite{park2020deformable,pumarola2021d,gafni2021dynamic,tretschk2021non}, relighting~\cite{boss2021nerd,srinivasan2021nerv}, generative models~\cite{schwarz2020graf,niemeyer2021giraffe,chan2021pi,jang2021codenerf}, etc. \wangcanc{Furthermore, DietNeRF~\cite{jain2021putting} designs a CLIP semantic consistency loss to improve few-shot NeRF and presents impressive results, and GRAF~\cite{schwarz2020graf} first adopts shape and appearance codes to conditionally synthesize NeRF, which inspires our adversarial
training.}

Despite the above success, a 3D model with NeRF representation is very unintuitive and difficult to edit since it is represented by millions of network parameters. To address this problem, the pioneering work EditNeRF~\cite{liu2021editing} defines a conditional NeRF, where the 3D object encoded by NeRF is conditioned on a shape code and an appearance code. By optimizing the adjustment to these two latent codes, user edits on shape and appearance color can be achieved. However, this method has limited capacity in shape manipulation as it only supports adding or removing local parts of the object. Also, the editing process of EditNeRF~\cite{liu2021editing} is slow because of its iterative optimization nature. Compared to EditNeRF~\cite{liu2021editing}, our method is different in three aspects. First, our method gives more freedom in shape manipulation and supports global deformation. Second, by learning two feed-forward networks mapping user edits to the latent codes, our method allows fast inference for the interactive editing. Moreover, different from the user scribbles used in EditNeRF~\cite{liu2021editing}, we introduce two intuitive ways to NeRF editing: using either a short
text prompt or an exemplar image, which are more friendly to novice users.

\noindent\textbf{CLIP-Driven Image Generation and Manipulation.}
An important building block of our work is CLIP~\cite{radford2021learning} which connects texts and images by bringing them closer in a shared latent space, under a contrastive learning manner. Powered by the CLIP model, some text-driven image generation and manipulation methods are proposed. Perez~\cite{perez2021imagesfromprompts} combines CLIP and StyleGAN~\cite{karras2020analyzing,karras2019style} to synthesize images by optimizing the latent code of a pre-trained StyleGAN according to a textual condition defined in the CLIP space. Instead of generating images from scratch, StyleCLIP~\cite{patashnik2021styleclip} introduces a text-based interface for StyleGAN to allow manipulations of real images with text prompts. Besides applying CLIP to GAN models,
DiffusionCLIP~\cite{kim2021diffusionclip} combines a diffusion model~\cite{song2019generative} with CLIP to conduct a text-driven image manipulation. It achieves a comparable performance to that of GAN-based image manipulation methods, with
the advantage of great mode coverage and training stability. However, all these methods only explore the text-guidance ability of CLIP, whereas our method unifies both text-and-image driven manipulations in a single model by fully exploiting the power of CLIP. Further, these methods are limited to image manipulation and fail to encourage multi-view consistency due to the lack of 3D information. In contrast, our model combines NeRF with CLIP, thus allowing editing 3D models in a view consistent way.


\begin{figure*}[tb]
\centering
\includegraphics[width=0.95\linewidth]{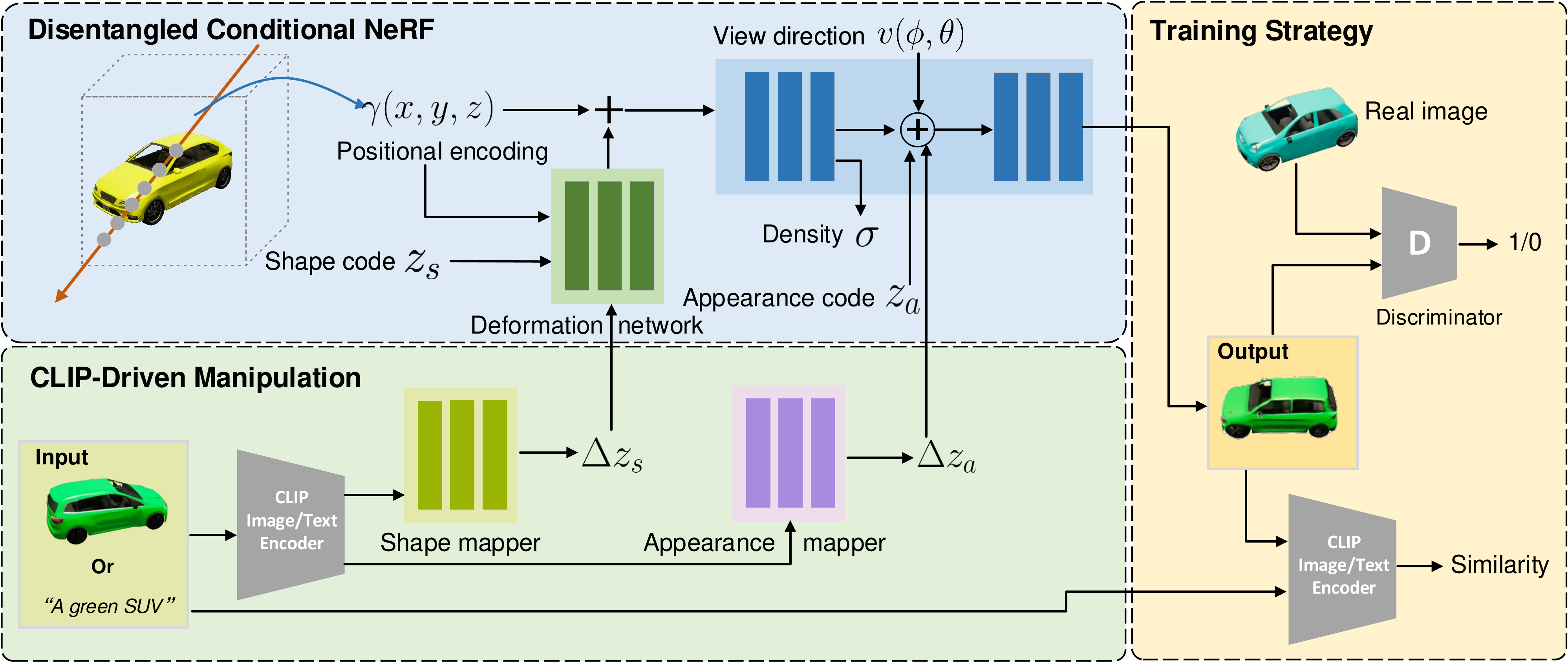}
   \caption{\textbf{The framework of the proposed method.} Our model first learns a disentangled conditional NeRF which takes positional encoding, view direction, shape code, and appearance code as input and outputs rendered image, while the shape code aims to deform the volume filed via a deformation network. This disentangled conditional NeRF is trained in an adversarial manner. Then given a reference image or a text prompt, the CLIP image or text encoder extracts the corresponding feature embedding for the shape and appearance mappers to lean a local step in the latent space for shape and appearance manipulation, respectively. These two mappers are trained using a CLIP similarity loss with our pre-trained disentangled conditional NeRF. }
   \label{fig:framework}
\end{figure*}

\section{Method}

\newcommand{\shapecode}{\vz_{s}}
\newcommand{\appearcode}{\vz_{a}}
\newcommand{\noise}{\vz_{n}}
\newcommand{\view}{\vv}
\newcommand{\network}{\mathcal{F}_\theta}
\newcommand{\shapemapper}{\mathcal{M}_{s}}
\newcommand{\appearmapper}{\mathcal{M}_{a}}
\newcommand{\mapper}{\mathcal{M}}
\newcommand{\deformnetwork}{\mathcal{T}}

In this section, we start with the general formulation of conditional NeRF (\S~\ref{subsec:cnerf}) as a 3D generative model conditioned by shape and appearance codes. We then present our disentangled conditional NeRF model (\S~\ref{subsec:disentangled}), which is able to individually control the shape and appearance manipulation. 
Next, we introduce our framework on leveraging the multi-modal power of CLIP for driving NeRF manipulation (\S~\ref{subsec:CDM}) using both text prompts or image exemplars, and the training strategy (\S~\ref{subsec:training}). Finally, we propose an inversion method (\S~\ref{subsec:inverse}) to allow editing a real image by a novel latent optimization approach on shape and appearance codes.

\subsection{Conditional NeRF}
\label{subsec:cnerf}
Built upon the original per-scene NeRF, conditional NeRF servers as a generative model for a particular object category, conditioned on the latent vectors that dedicatedly control shape and appearance. Specifically, conditional NeRF is represented as a continuous volumetric function $\network$ that maps a 5D coordinate (a spatial position $\vx(x,y,z)$ and a view direction $\view(\phi,\theta)$), together with a shape code $\shapecode$ and an appearance code $\appearcode$, to a volumetric density $\sigma$ and a view-dependent radiance $\vc(r,g,b)$, parametrized by a multi-layer perceptron (MLP). A trivial formulation $\mathcal{F}'_\theta(\cdot)$ of conditional NeRF can be:
\begin{equation}
\mathcal{F}'_\theta(\vx,\view,\shapecode,\appearcode):\big(\mathit{\Gamma}(\vx)\oplus\shapecode,\mathit{\Gamma}(\view)\oplus\appearcode\big)\rightarrow(\vc,\sigma),
\label{eq:contional_nerf_1}
\end{equation}
where $\oplus$ is the concatenation operator.

Here $\mathit{\Gamma}(\vp)=\big\{\gamma(p)\mid p\in\vp\big\}$ is the sinusoidal positional encoding that separately projects each coordinate $p$ of vector $\vp$ to a high dimensional space. Each output dimension of the encoding function $\gamma(\cdot):\mathbb{R}\rightarrow\mathbb{R}^{2m}$ is defined as:
\begin{equation}
\gamma(p)_k=
\begin{cases} 
\sin(2^k\pi p),&\mbox{if }k\mbox{ is even},\\
\cos(2^k\pi p),&\mbox{if }k\mbox{ is odd},
\end{cases}
\label{eq:positional_encoding}
\end{equation}
where $k\in\{0,\ldots,2m-1\}$ and $m$ is a hyper-parameter that controls the total number of frequency bands.

\subsection{Disentangled Conditional NeRF}
\label{subsec:disentangled}
The aforementioned conditional NeRF does introduce conditional generation capability to the NeRF architecture. However, this trivial formulation $\mathcal{F}'_\theta$ (Eq.~\ref{eq:contional_nerf_1}) suffers from mutual intervention between shape and appearance conditions, e.g., manipulating the shape code could also cause color changes.
In observation of this issue, we propose our disentangled conditional NeRF architecture to achieve individual control over both shape and appearance by properly disentangling the conditioning mechanism.

\textbf{Conditional Shape Deformation.}
Rather than directly concatenating the latent shape code to the encoded position feature, we propose to formulate the shape conditioning through explicit volumetric deformation to the input position. This conditional shape deformation not only improves the robustness of the manipulation and preserves the original shape details as much as possible by regularizing the output shape to be a smooth deformation of the base shape, but more importantly also completely isolates the shape condition from affecting the appearance.

To this end, we design a shape deformation network $\deformnetwork:(\vx,\shapecode)\rightarrow\Delta\vx$, which projects a position $\vx$ and the input $\shapecode$ to displacement vectors $\Delta\vx\in\mathbb{R}^{3\times2m}$ corresponding to each band of the positional encoding $\mathit{\Gamma}(\vx)$. Thus, the deformed positional encoding $\mathit{\Gamma}^*(\vp,\shapecode)=\{\gamma^*(p,\Delta p)\mid p\in\vp,\Delta p\in\deformnetwork(\vp,\shapecode)\}$ is defined as:
\begin{equation}
\gamma^*(p,\Delta p)_k=\gamma(p)_k+\tanh(\Delta p_k),
\label{eq:deformed_positional_encoding}
\end{equation}
where the scalar $p$ and the vector $\Delta p\in\mathbb{R}^{2m}$ belong to the same axis from $\vp$ and $\Delta\vp$. The hyperbolic tangent function $\tanh(\cdot)$ is used to constrain the displacements in the range of $[-1,1]$, which helps avoid poor local minimums caused by large motions and increase the training robustness.
\wangcanc{Our deformation network is inspired by Nerfies~\cite{park2020deformable} but has the major difference as:  the deformation network in Nerfies takes a latent deformation code to deform the local volume field of a single fixed object or scene, while ours uses a shape code to deform the global volume field of various object instances that can be generated by the conditional NeRF.}

\textbf{Deferred Appearance Conditioning.}
In NeRF, the density is predicted first as a function of position and the radiance is then predicted from both position and view direction. Similar to Graf~\cite{schwarz2020graf} and EditNeRF~\cite{liu2021editing}, we also defer the appearance conditioning to concatenate the appearance code with the view direction as the input to the radiance prediction network, which allows manipulating the appearance without touching the shape information, i.e., density.

Overall, as illustrated in Fig.~\ref{fig:framework}, our disentangled conditional NeRF $\mathcal{F}_\theta(\cdot)$ is defined as:
\begin{equation}
\mathcal{F}_\theta(\vx,\view,\shapecode,\appearcode):\big(\mathit{\Gamma}^*(\vx,\shapecode),\mathit{\Gamma}(\view)\oplus\appearcode\big)\rightarrow(\vc,\sigma).
\label{eq:contional_nerf_2}
\end{equation}
And for the simplicity of notation, we use $\mathcal{F}_\theta(\view,\shapecode,\appearcode)=\big\{\mathcal{F}_\theta(\vx,\view,\shapecode,\appearcode)\mid\vx\in\mathbf{R}\big\}$ to denote the rendering of the whole image with viewport $\mathbf{R}$.

\begin{figure}[tb]
\centering
\includegraphics[width=0.98\linewidth]{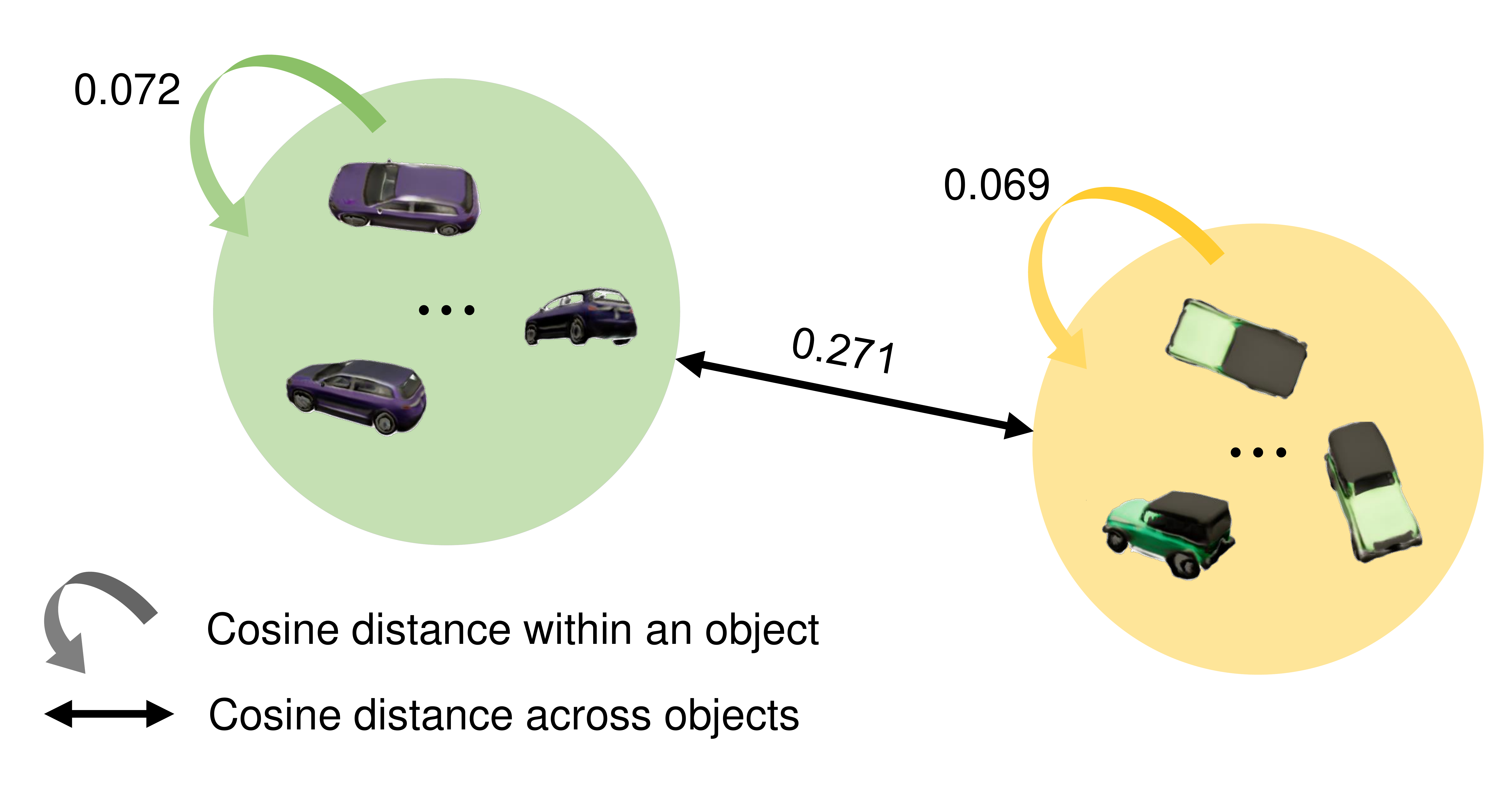}
  \caption{\textbf{Multi-View Consistency Evaluation of CLIP.} We randomly select two cars and measure their pairwise CLIP cosine distances of 1) different cars in a same view and 2) a same car in different random views (we sampled 144 views from the upper hemisphere, as a combination of 12 $\phi$ and 12 $\theta$ poses). Though the camera poses vary dramatically, different views for a same object have higher similarity (small distance). But different objects have lower similarity (large distance) even in an identical view.}
  \label{fig:clip_evalue}
\end{figure}

\subsection{CLIP-Driven Manipulation}
\label{subsec:CDM}
With our disentangled conditional NeRF (Eq.~\ref{eq:contional_nerf_2}) as a generator, we now introduce how we integrate the CLIP model into the pipeline to achieve text-driven manipulation on both shape and appearance.

To avoid optimizing both shape and appearance codes for each target sample, which tends to be versatile and time-consuming, we take a feed-forward approach to directly update the condition codes from the input text prompt. Specifically, given an input text prompt of $\vt$ and the initial shape/appearance code of $\shapecode'$/$\appearcode'$, we train a shape mapper $\shapemapper$ and an appearance mapper $\appearmapper$ to update the codes as:
\begin{equation}
\begin{split}
\shapecode=\shapemapper\big(\hat{\mathcal{E}}_t(\vt)\big)+\shapecode',\\
\appearcode=\appearmapper\big(\hat{\mathcal{E}}_t(\vt)\big)+\appearcode',
\end{split}
\label{eq:clip-mapper}
\end{equation}
where $\hat{\mathcal{E}}_t(\cdot)$ is the pre-trained CLIP text encoder that projects the text to the CLIP embedded feature space and both mappers map this CLIP embedding to displacement vectors that update the original shape and appearance codes.

In addition, given that CLIP includes an image encoder and a text encoder mapping to a joint embedding space, we define a cross-modal CLIP distance function $D_\text{CLIP}(\cdot,\cdot)$ to measure the embedding similarity between the input text and a rendered image patch:
\begin{equation}
\label{eq:clip-loss}
    D_{\text{CLIP}}(\mathbf{I},\vt)=1-\big\langle\hat{\mathcal{E}}_i(\mathbf{I}),\hat{\mathcal{E}}_t(\vt)\big\rangle,
\end{equation} 
where $\hat{\mathcal{E}}_i(\cdot)$ and $\hat{\mathcal{E}}_t(\cdot)$ are the pre-trained CLIP image and text encoders, $\mathbf{I}$ and $\vt$ are the input image patch and text, and $\langle\cdot,\cdot\rangle$ is the cosine similarity operator.

Without loss of generality, here we assume that the manipulation control comes from a text prompt $\vt$. However, our distance can also be extended to measure similarity between two images or two text prompts. Thus, our framework naturally supports editing with an image exemplar by trivially replacing the text prompt with this exemplar in aforementioned equations.

\noindent\textbf{Discussion.}
To perform NeRF manipulation with image-level CLIP model, a natural question is whether the CLIP feature is stable across different viewpoints and whether it can distinguish object differences.
To evaluate this, we randomly select two objects (e.g., an SUV and a jeep) and measure the pairwise CLIP-space cosine distances between 1) different views of a same object, and 2) different objects in a same view. As shown in \Fref{fig:clip_evalue}, we find the distance is more sensitive to small object difference than large view variations.
This suggests that a pre-trained CLIP model has the ability to support view-consistency representations for 3D-aware applications. 
\wangcanc{A similar observation is found by DietNeRF~\cite{jain2021putting} and applied in 3D reconstruction.}

\subsection{Training Strategy}
\label{subsec:training}
Our pipeline is trained in two stages: we first train the disentangled conditional NeRF including the conditional NeRF generator and the deformation network; then we fix the weights of the generator and train the CLIP manipulation parts including both the shape and appearance mappers.

\textbf{Disentangled Conditional NeRF.}
Our conditional NeRF generator $\network$ is trained together with the deformation network using a non-saturating GAN objective~\cite{mescheder2018training} with the discriminator $\mathcal{D}$, where $f(x)=-\log\big(1+\exp(-x)\big)$ and $\lambda_r$ is the regularization weight. Assuming that real images $\mathbf{I}$ form the training data distribution of $d$, we randomly sample the shape code $\shapecode$, the appearance code $\appearcode$, and the camera pose from $\mathcal{Z}_s$, $\mathcal{Z}_a$, and $\mathcal{Z}_v$, respectively, where $\mathcal{Z}_s$ and $\mathcal{Z}_a$ are the normal distribution, and $\mathcal{Z}_v$ is the upper hemisphere of the camera coordinate system.
\begin{equation}
\begin{split}
    \mathcal{L}_\text{GAN}=&\mathbb{E}_{\shapecode\sim\mathcal{Z}_s,\appearcode\sim\mathcal{Z}_a,\view\sim\mathcal{Z}_v}\big[f\big(\mathcal{D}(\network(\view,\shapecode,\appearcode))\big)\big]\\
    +&\mathbb{E}_{\mathbf{I}\sim d}\big[f\big(-\mathcal{D}(\mathbf{I})+\lambda_r\|\nabla\mathcal{D}(\mathbf{I})\|^2\big)\big].
\end{split}
\end{equation}

\textbf{CLIP Manipulation Mappers.}
We use pre-trained NeRF generator $\network$, CLIP encoders $\{\hat{\mathcal{E}}_t,\hat{\mathcal{E}}_i\}$, and the discriminator $\mathcal{D}$ to train the CLIP shape mapper $\shapemapper$ and appearance mapper $\appearmapper$. All network weights, except the mappers, are fixed, denoted as $\{\hat{\cdot}\}$. Similar to the first stage, we randomly sample the shape code $\shapecode$, the appearance code $\appearcode$, and the camera pose $\view$ from their respective distributions. In addition, we sample the text prompt $\vt$ from a pre-defined text library $\mathbf{T}$. By using our CLIP distance $D_\text{CLIP}$ (Eq.~\ref{eq:clip-loss}) with weight $\lambda_c$, we train the mappers with the following losses:
\begin{equation}
\begin{split}
    \mathcal{L}_\text{shape}=f\big(\hat{\mathcal{D}}\big(&\hat{\network}\big(\view,\shapemapper(\hat{\mathcal{E}}_t(\vt))+\shapecode,\appearcode\big)\big)\big)+\\
    \lambda_c D_\text{CLIP}\big(&\hat{\network}\big(\view,\shapemapper(\hat{\mathcal{E}}_t(\vt))+\shapecode,\appearcode\big),\vt\big),
\end{split}
\end{equation}
\begin{equation}
\begin{split}
    \mathcal{L}_\text{appear}=f\big(\hat{\mathcal{D}}\big(&\hat{\network}\big(\view,\shapecode,\appearmapper(\hat{\mathcal{E}}_t(\vt))+\appearcode\big)\big)\big)+\\
    \lambda_c D_\text{CLIP}\big(&\hat{\network}\big(\view,\shapecode,\appearmapper(\hat{\mathcal{E}}_t(\vt))+\appearcode\big),\vt\big).
\end{split}
\end{equation}

\subsection{Inverse Manipulation}
\label{subsec:inverse}
The manipulation pipeline we have introduced so far works on an initial sample with known conditions including the shape and appearance codes. To apply the manipulation to an input image $\mathbf{I}_r$ belonging to the same training category, the key is to first optimize all generation conditions to inversely project the image to the generation manifold, similar to the latent image manipulation methods~\cite{abdal2019image2stylegan,abdal2020image2stylegan++,gu2020image,pan2021exploiting}. Following the EM algorithm~\cite{dempster1977maximum}, we design an iterative method to alternatively optimize the shape code $\shapecode$, the appearance code $\appearcode$, and the camera $\view$.

To be specific, during each iteration, we first optimize $\view$ while keeping $\shapecode$ and $\appearcode$ fixed using the following loss:
\begin{equation}
\label{eq:pose-loss}
\begin{split}
\mathcal{L}_v=\big\|\hat{\network}(\view,\hat{\shapecode},\hat{\appearcode})-\mathbf{I}_r\big\|_2+\\
\lambda_v D_{\text{CLIP}}\big(\hat{\network}(\view,\hat{\shapecode},\hat{\appearcode}),\mathbf{I}_r\big).
\end{split}
\end{equation}

We then update the shape code by minimizing:
\begin{equation}
\label{eq:shape-loss}
\begin{split}
\mathcal{L}_s=\big\|\hat{\network}(\hat{\view},\shapecode+\lambda_n\vz_n,\hat{\appearcode})-\mathbf{I}_r\big\|_2+\\
\lambda_s D_{\text{CLIP}}\big(\hat{\network}(\hat{\view},\shapecode+\lambda_n\vz_n,\hat{\appearcode}),\mathbf{I}_r\big),
\end{split}
\end{equation}
where $\appearcode$ and $\view$ are fixed, $\noise$ is a random standard Gaussian noise vector sampled in each step to improve the optimization robustness, and $\lambda_{n}$ linearly decays from $1$ to $0$ through the whole optimization iterations.

The appearance code is updated in a similar manner:
\begin{equation}
\label{eq:appearance-loss}
\begin{split}
\mathcal{L}_a=\big\|\hat{\network}(\hat{\view},\hat{\shapecode},\appearcode+\lambda_n\vz_n)-\mathbf{I}_r\big\|_2+\\
\lambda_a D_{\text{CLIP}}\big(\hat{\network}(\hat{\view},\hat{\shapecode},\appearcode+\lambda_n\vz_n),\mathbf{I}_r\big),
\end{split}
\end{equation}


\section{Experiments}

\begin{figure}[tb]
\centering
\includegraphics[width=1.0\linewidth]{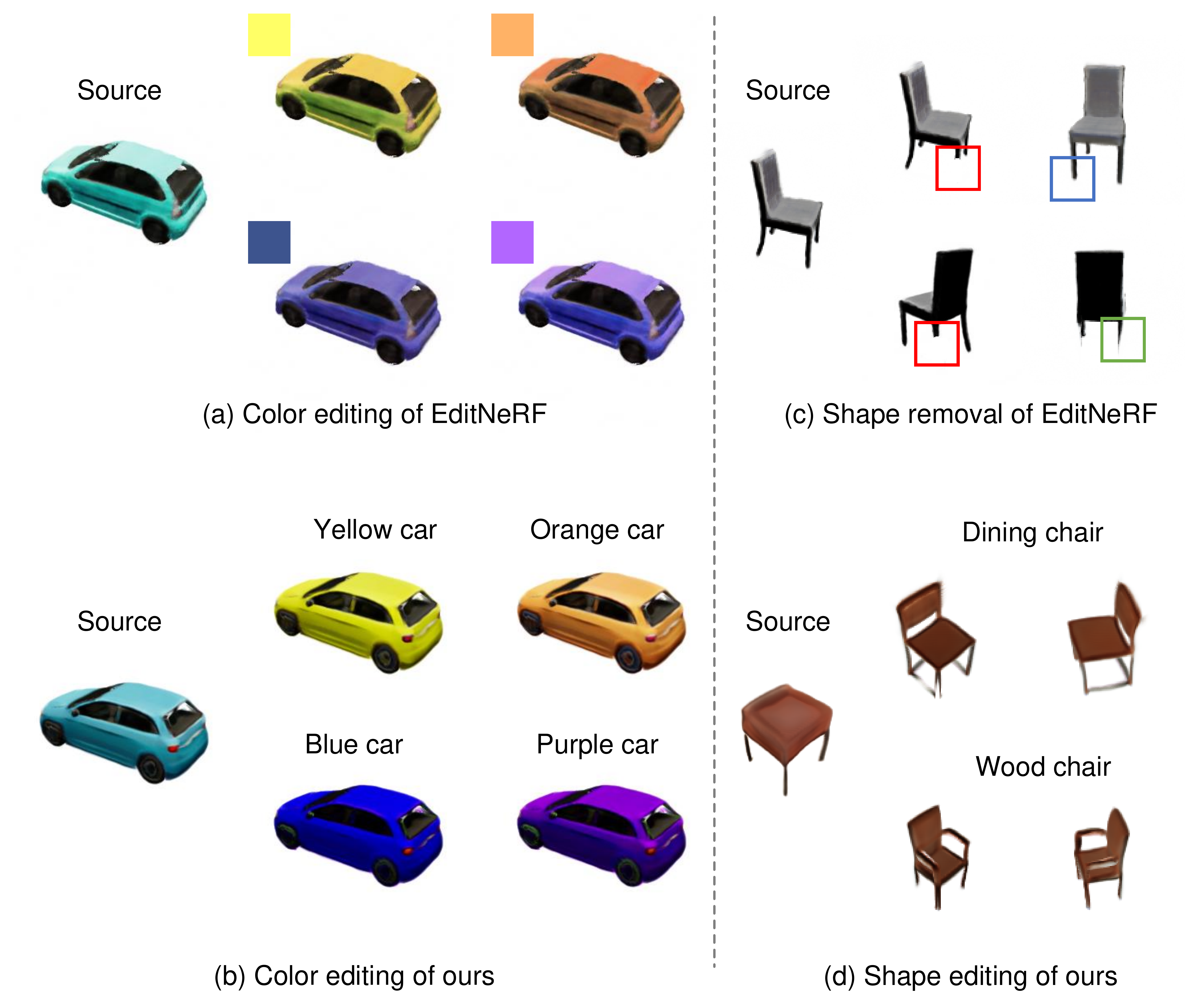}
  \caption{Compared to EditNeRF.}
  \label{fig:cmp_ednerf}
\end{figure}

\textbf{Datasets.} We evaluate our method on two public datasets: \textit{Photoshapes}~\cite{park2018photoshape,schwarz2020graf} with 150K chairs rendered at 128$\times$128 following the rendering
protocol of~\cite{oechsle2020learning} and \textit{Carla} with 10K cars rendered at 256$\times$256 using the Driving simulator~\cite{dosovitskiy2017carla,schwarz2020graf}. Each object is rendered in a random view without providing any camera pose parameters.

\textbf{Implementation details.} Our conditional NeRF is an 8-layer MLP with each layer containing 256 hidden units, and the input dimension is 64. Following the default architecture for NeRF ~\cite{mildenhall2020nerf}, we also use ReLU activations. The deformation network is a 4-layer MLP with ReLU activations and 256 hidden units per layer. It takes a 128-dimensional shape code $\shapecode$ as input, $\shapecode\in\mathbb{R}^{128}$. We also represent the appearance code $\appearcode$ using 128 dimensions, $\appearcode\in\mathbb{R}^{128}$. Both shape and appearance mappers are 2-layer MLPs with ReLU activations. The channel sizes of each mapper are 128, 256, and 128, respectively. The implementation of the discriminator follows PatchGAN~\cite{isola2017image}. We use the Adam optimizer and an initial learning rate of $10^{-4}$ to train the network. The learning rate is decayed by 0.5 every 50K steps. In the inversion, we also use the Adam optimizer with the learning rate starting from $10^{-3}$ and decreasing by 0.75 every 100 steps. Besides,
$\lambda_r=0.5$, $\lambda_v= 0.1$, and $\lambda_s=\lambda_a=0.2$.
All the models are trained on an NVIDIA V100 GPU platform.

\begin{table}[t]
\tabcolsep=4.0pt
\begin{tabular}{lcccccc}
    \toprule
         & \multicolumn{2}{c}{Chairs} & \multicolumn{2}{c}{Cars} \\
    \cmidrule(r){2-3}\cmidrule(r){4-5}
      & Shape & Appearance & Shape & appearance \\
    \midrule
    EditNeRF & 30.0  & 15.9 & 33.2 & 16.8 \\	 
    Ours & 0.58 & 0.51 & 2.12 & 1.98 \\
    \bottomrule
\end{tabular}
\caption{\textbf{Compared to EditNeRF~\cite{liu2021editing} on editing time averaged on 20 images.} We only include the inference/optimization time(s) and single-view rendered time(s) for chairs~(128$\times$128 pixels) and cars~(256$\times$256 pixels).
}
\label{tbl:cmp_time}
\end{table}

\begin{figure}[tb]
\centering
\includegraphics[width=1.0\linewidth]{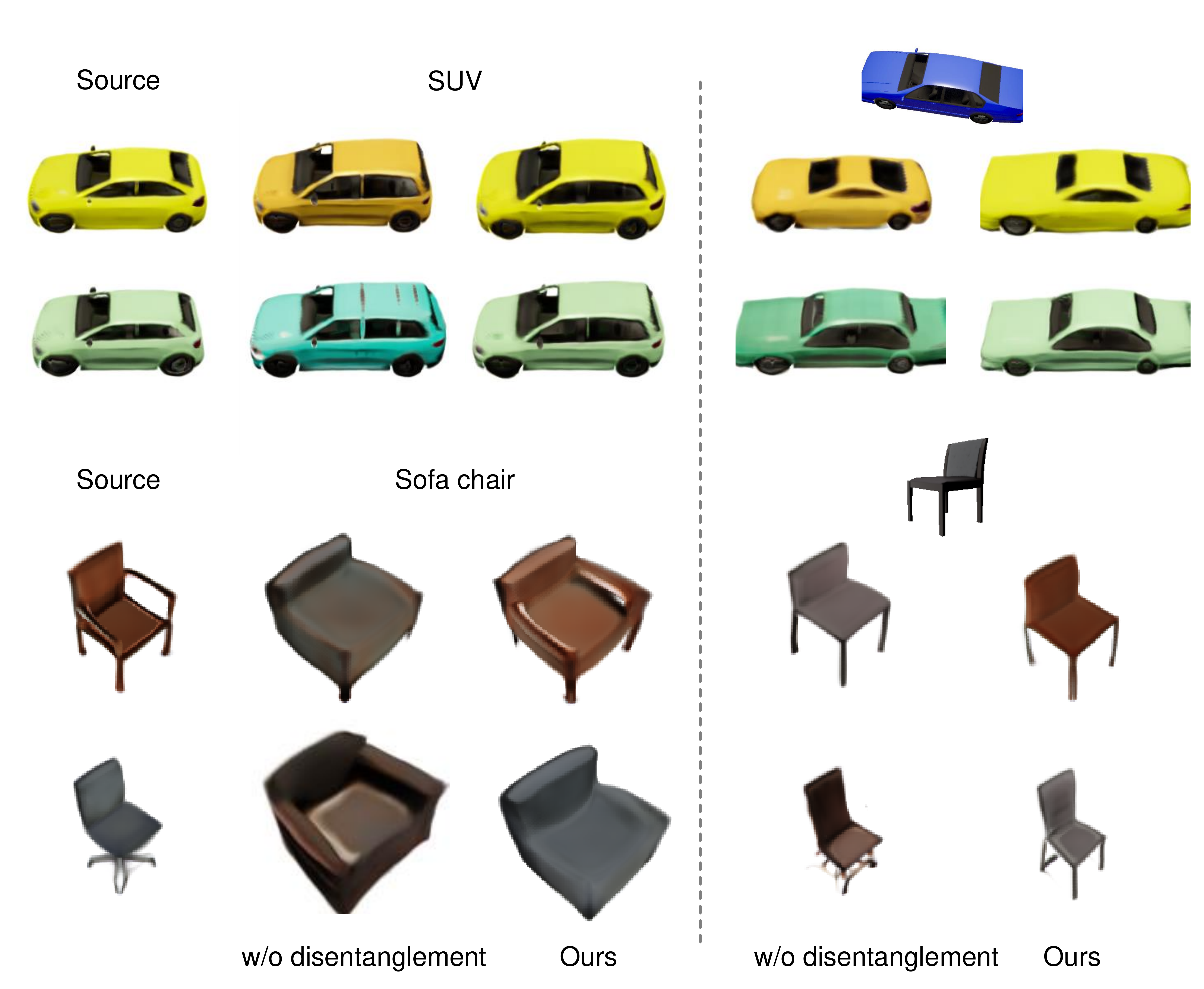}
  \caption{\textbf{Ablation Study for Disentanglement.} We show text-and-exemplar driven shape editing results of our method and the baseline method without using our disentangled technique. When editing the shape, the latter can change the appearance, while ours keeps the appearance unchanged.}
  \label{fig:ablation_disentangle}
\end{figure}

\subsection{Compared to EditNeRF}

We compare with pioneering work in NeRF editing, EditNeRF~\cite{liu2021editing} on the editing of shape and appearance color of both datasets in \Fref{fig:cmp_ednerf}. For the Photoshapes dataset, EditNeRF is trained using 600 instances with 40 views per instance while ours uses only one view. For the Carla dataset, EditNeRF uses 10K cars with a single view per instance, same as ours. Besides, camera pose parameters are required during training of EditNeRF but unknown for us.

We first compare the capability and performance between EditNeRF and our method. For the color editing~(\Fref{fig:cmp_ednerf}-(a)), EditNeRF requires the user to select a target color and draw coarse scribbles on a local region. With the foreground and background masks created by the coarse scribbles, EditNeRF performs the appearance editing by optimizing the appearance code and a conditional NeRF to achieve the target color. We observe that unnatural color effects appear on the edited results of EditNeRF (\textit{e.g.} discontinuity on car doors), and the generated color is not completely faithful to the target color. In contrast, we allow the user to change the color more simply by providing a text prompt and our method produces more natural editing results~(\Fref{fig:cmp_ednerf}-(b)). For the shaping editing~(\Fref{fig:cmp_ednerf}-(c)), EditNeRF can only support local shape editing, such as shape part removal. Given the user's editing scribbles which for example indicate to remove a leg of a chair (in the red rectangles), EditNeRF optimizes a few layers in the network to fit the shape in the input view but it cannot ensure successful propagation to unseen views (in the blue rectangle) and keep the structure of other parts intact (in the green rectangle). Compared to it, our method supports a large degree of shape deformation and generalizes well to unseen views~(\Fref{fig:cmp_ednerf}-(d)). Besides, EditNeRF, as an optimized-based method, takes a large amount of time for the optimization, while our feedforward code mappers achieve much faster inference of the target shape and appearance~(\Tref{tbl:cmp_time}).

To quantitatively evaluate how good the image quality is preserved after editing, we calculate the FID scores of 2K testing images before and after editing. Due to training with 40 views per instance, EditNeRF shows better reconstruction before editing on the chair dataset, but its editing notably degrades the image quality while our method ensures comparable quality before and after manipulation. On the car dataset, the performance of EditNeRF significantly drops because of only one view per instance used in training. Trained under the same setting, our model improves reconstruction quality by a large margin and well preserves the quality during editing. \lj{Since EditNeRF requires user scribbles for shape editing and is difficult to generate a large set of results with random conditions}, we exclude it in the comparison on shape editing while our method performs equally well regardless of editing shape or color.

\begin{figure}[tb]
\centering
\includegraphics[width=0.85\linewidth]{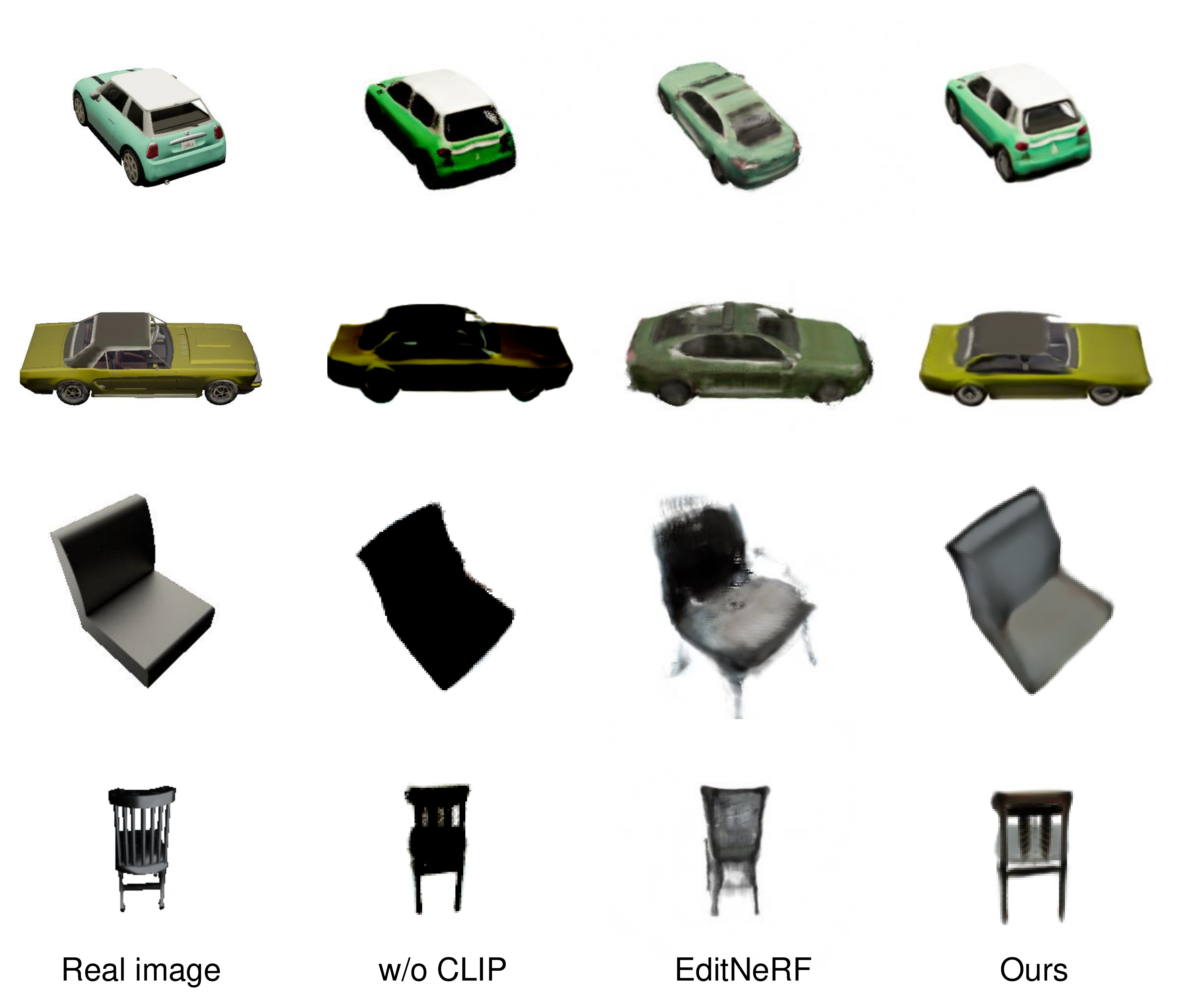}
  \caption{\textbf{Ablation study on our inversion method and comparison with EditNeRF.}}
  \label{fig:cmp_inversion}
\end{figure}

We also compare with EditNeRF in the inversion results (\Fref{fig:cmp_inversion}). EditNeRF infers the shape and appearance codes by fine-tuning the condition NeRF using the standard NeRF photometric loss. Our optimized-based inversion method outperforms by benefiting from CLIP's ability to provide multi-view consistency representations~(more discussions in \sref{subsec:CDM} and ablation in \ref{subsec:AS}).

\begin{figure*}[tb]
\centering
\includegraphics[width=1.0\linewidth]{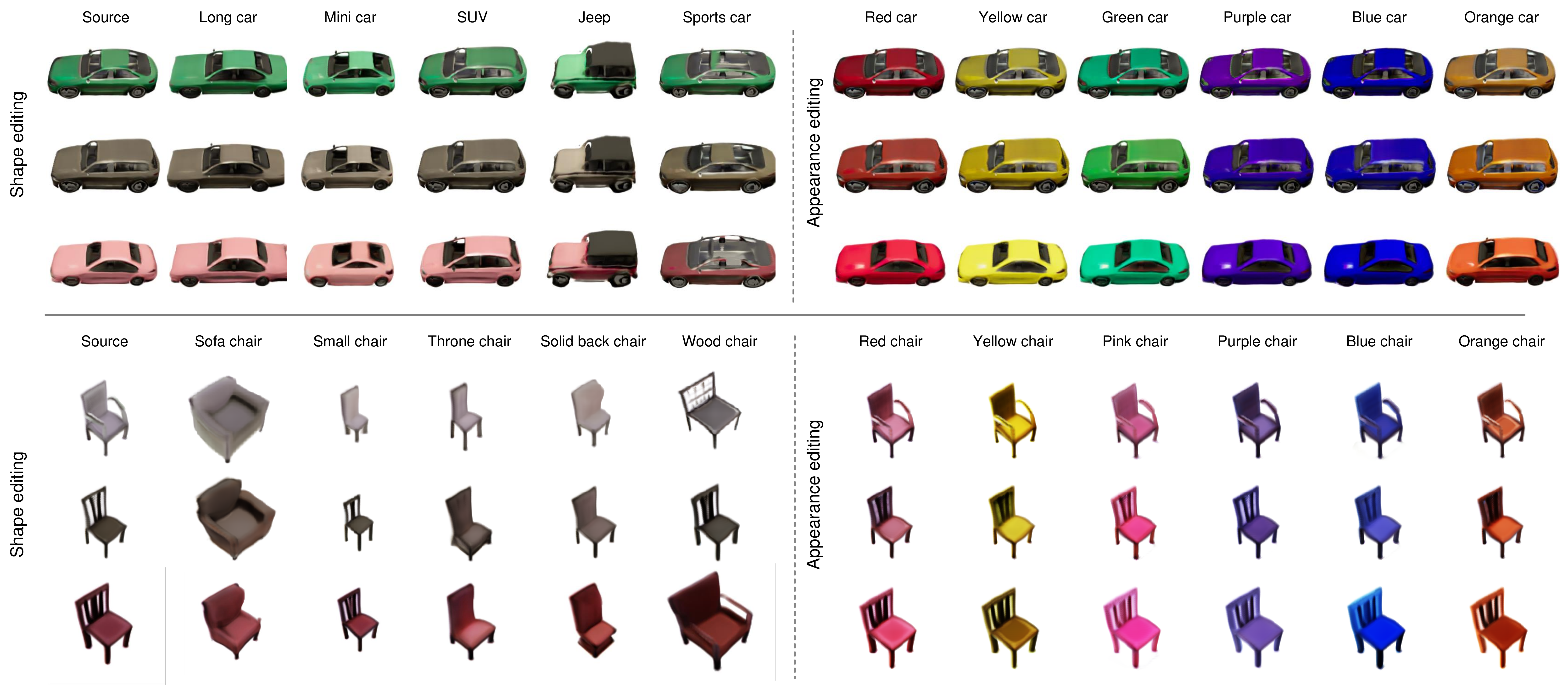}
   \caption{\textbf{Text-Driven Editing Results.}}
   \label{fig:text_results}
\end{figure*}

\subsection{Ablation Study}
\label{subsec:AS}
\begin{table}[t]
\tabcolsep=2.0pt
\begin{tabular}{clcccccccc}
    \toprule
      &   & \multicolumn{3}{c}{Chairs} & \multicolumn{3}{c}{Cars} \\
    \cmidrule(r){3-5}\cmidrule(r){6-8}
     & & Before & After & Diff. & Before & After & Diff. \\
    \midrule
    \multirow{3}{*}{(a)}  & EditNeRF  & 36.8  & 40.2 & 3.4 & 102.8 & 118.7 & 15.9 \\	 
    & w/o disen. &  52.5 & 54.3 & 1.8 & 69.2 & 69.9 & 0.7 \\
    & Ours & 47.8 & 49.0 & 1.2 & 66.7 & 67.2 & 0.5 \\
    \midrule
    \multirow{2}{*}{(b)} & w/o disen. &  52.5 & 53.2 & 0.7 & 69.2 & 71.1 & 1.9  \\
    & Ours & 47.8 & 48.4 & 0.6 & 66.7 & 67.8 & 1.1 \\
    \bottomrule
\end{tabular}
\caption{\textbf{Fr\'echet inception distance (FID) for evaluating the image quality of reconstructed views before and after editing on: (a) color and (b) shape (lower value means better).} We use 2K images with various views drawn randomly from the latent space to calculate the FIDs for reconstructed images, and then perform various edits on these images to recalculate FIDs of edited results. As EditNeRF requires user scribbles for shape editing, we exclude it from the comparison on shape manipulation. w/o disen. is a variant of our model without the shape deformation network used for disentangling control of shape and appearance.}

\label{tbl:FID}
\end{table}

We evaluate our model w/ and w/o the disentangled design (\sref{subsec:disentangled}). In \Fref{fig:ablation_disentangle}, the model trained without the conditional shape deformation network (\textit{i.e.} w/o disen.) frequently introduces color changes when performing shape editing. In contrast, our disentangled conditional NeRF achieves individual shape control because the conditional shape deformation network is able to isolate the shape condition from the appearance control and deform the base volume field to generate new objects without affecting the appearance. Also, since the deformation network implicitly enforces regularization of the generated shape, the resulting quality is further improved as shown in~\Tref{subsec:AS}.

We conduct another ablation study in \Fref{fig:cmp_inversion} to evaluate our inversion optimization method. The baseline method~(w/o CLIP) only computes the standard NeRF photometric loss between the output and a single image. Its result quality is limited due to the difficulty in inferring a complete 3D NeRF model from a single view. As discussed in \sref{subsec:CDM}, CLIP has the capability to produce robust pose-invariant features. Therefore, our inversion method introduces a CLIP constraint during optimization and achieves better inversion results thanks to the CLIP prior.

\begin{figure*}[h]
\centering
\includegraphics[width=1.0\linewidth]{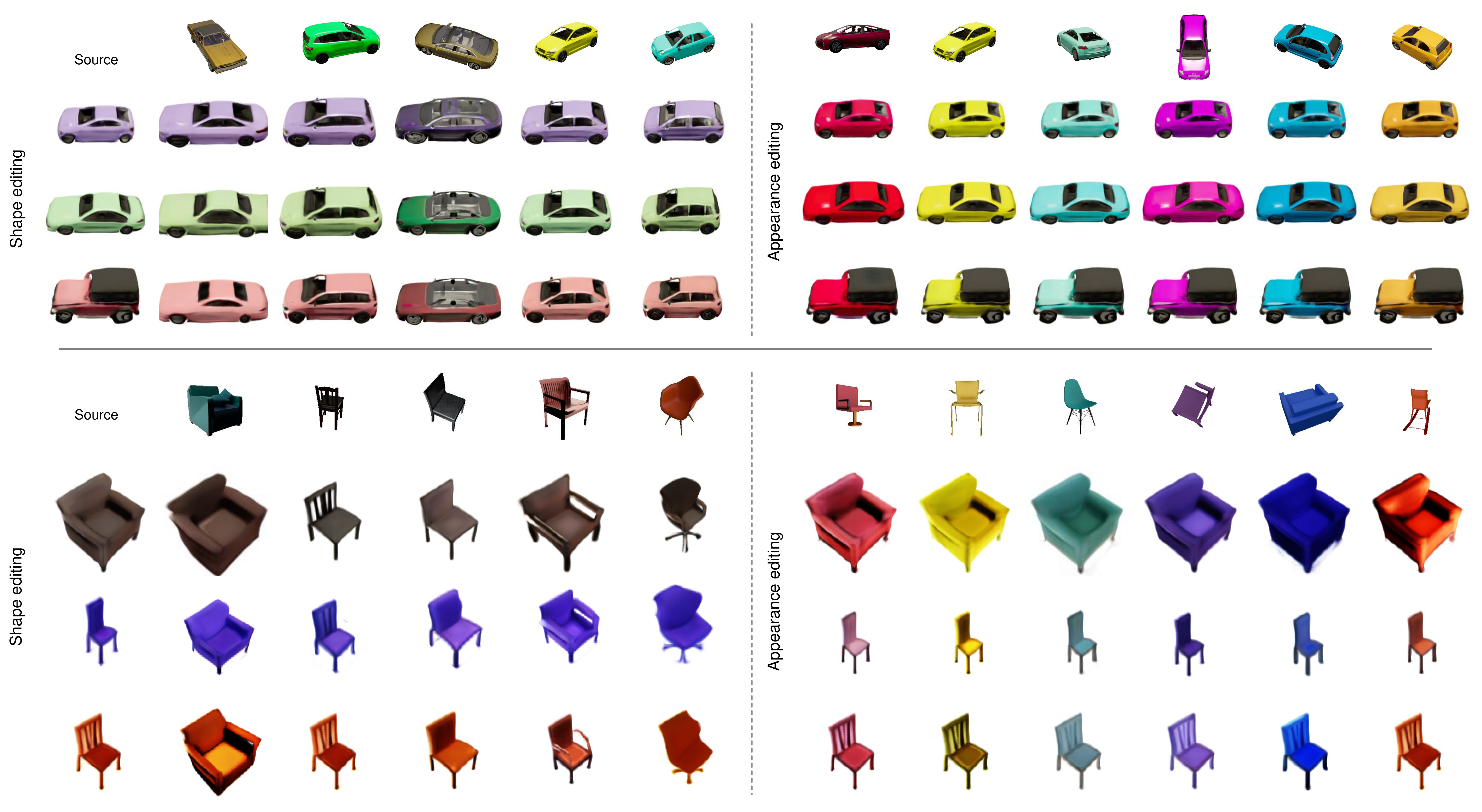}
   \caption{\textbf{Exemplar-Driven Editing Results.}}
   \label{fig:image_results}
\end{figure*}

\subsection{CLIP-Driven Manipulation}

Our method supports editing of object shape or appearance using text. When manipulating the shape, we keep the appearance code unchanged and the same applies to appearance editing. \Fref{fig:text_results} demonstrates diverse editing results. Note that when the car with a light color is deformed to a sports car, its color may become darker. But it is not a failure case, as the colors of all sports cars in the Carla dataset are inherently intenser. Besides, we find that our method naturally preserves the shading when changing the appearance color.
When editing the chair shape, if the user's input text is highly relevant to the source shape$-$for example, the source chair is a wood chair, and the user also wants a 'wood chair'$-$the result will be slightly different from the source. During the color editing, our method guarantees that the shape is completely preserved. 
Our method also supports exemplar-based manipulation by providing a real target image instead of a text prompt. 
We present various exemplar-guided shape and appearance editing results in \Fref{fig:image_results}.
Our method achieves semantic-precise and individual control of shape and appearance referring to the exemplar image.

\begin{figure}[tb]
\centering
\includegraphics[width=1.0\linewidth]{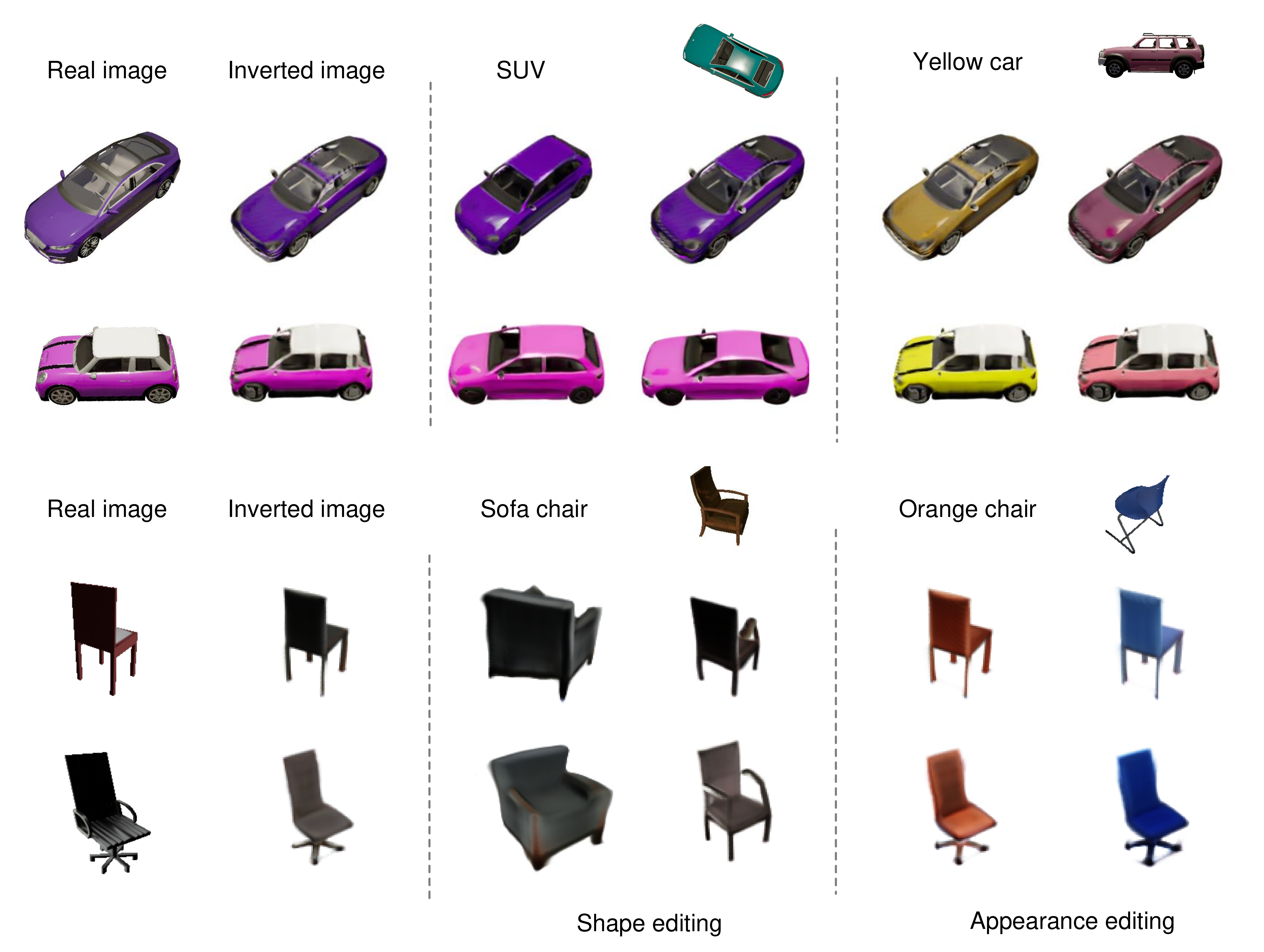}
  \caption{\textbf{Editing Results on Real Images.}}
  \label{fig:inverse_editing}
\end{figure}

\subsection{Real Image Manipulation}

To evaluate the generalization ability of our model in processing a single real image that does not exist in our training set, we experiment with the real image by inverting it to a shape code and an appearance code and then applying them to edit. We show the inverted and edited results in \Fref{fig:inverse_editing}. We observe that inverting the chair is much more challenging than inverting a car due to the chair's delicate structures, such as the wheels of the office chair. 
However, even the office chair is not perfectly reconstructed, the editing ability of our method is not affected. Our method still ensures accurate editing in shape and appearance.

\begin{table}[t]
\tabcolsep=5.2pt
\begin{tabular}{cccccc}
    \toprule
         \multicolumn{3}{c}{Chairs} & \multicolumn{3}{c}{Cars} \\
    \cmidrule(r){1-3}\cmidrule(r){4-6}
       Text & Exemplar & Avg. & Text & Exemplar & Avg.  \\
    \midrule
     0.821  & 0.877 & 0.849 & 0.814 & 0.859 & 0.837 \\
    \bottomrule
\end{tabular}
\caption{\textbf{User Study Results.} We report correct matching rate by counting whether an editing result is matched to the right text/exemplar guidance by users. 
}
\label{tbl:user_study}
\end{table}

\subsection{User Study}

We conduct a user study to evaluate the perceptual quality and accuracy of the editing results. We include 20 questions in the study, each question with 5 results of cars or chairs generated by 5 randomly selected text prompts or 5 randomly selected exemplars. We randomly shuffle the results and give users unlimited time to match each result with the correct text or image. We collect answers from 23 participates and report the matching accuracy rate in~\Tref{tbl:user_study}. Our method, in more than 80\% cases, succeeds in editing objects exactly corresponding to the description given by the text or the exemplar.


\begin{figure}[h]
\centering
\includegraphics[width=1.0\linewidth]{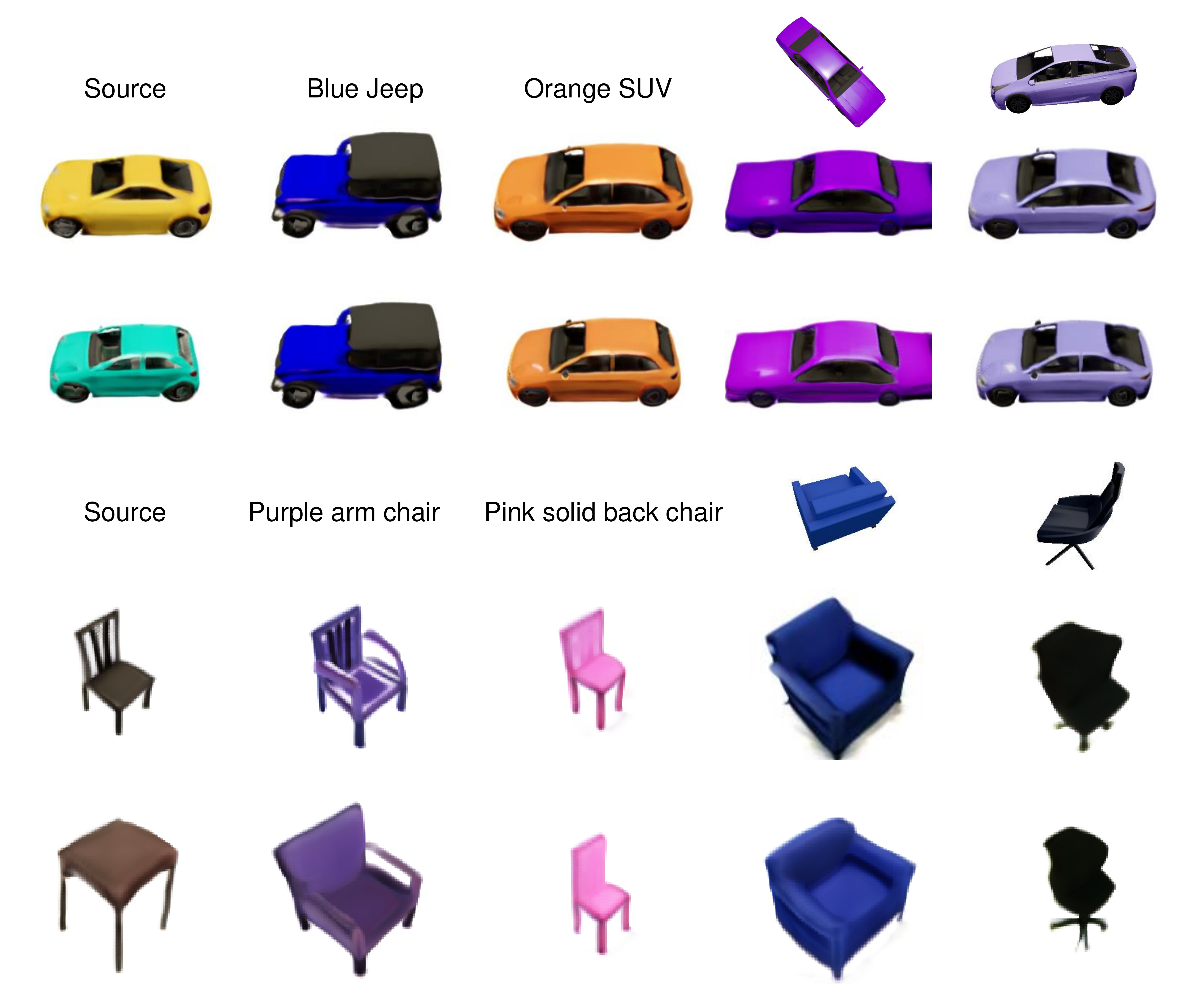}
  \caption{Continuous manipulation results. Our method supports editing both the shape and appearance with a single text prompt or an exemplar by continuously editing the shape and appearance.}
  \label{fig:extend_contine}
\end{figure}

\begin{figure}[h]
\centering
\includegraphics[width=1.0\linewidth]{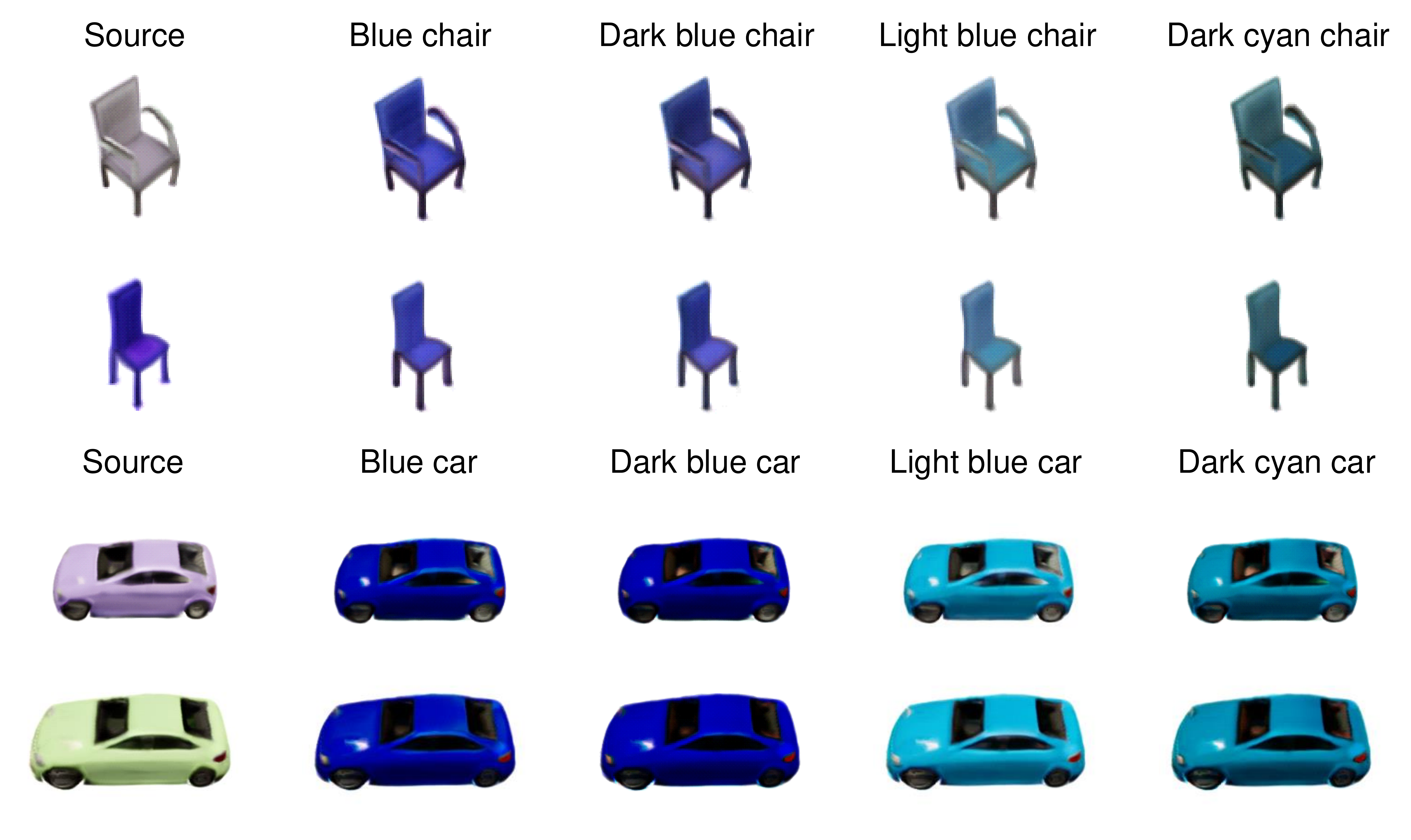}
  \caption{Fine-grained appearance manipulation results within a same color category.}
  \label{fig:extend_fine}
\end{figure}

\begin{figure}[t]
\centering
\includegraphics[width=1.0\linewidth]{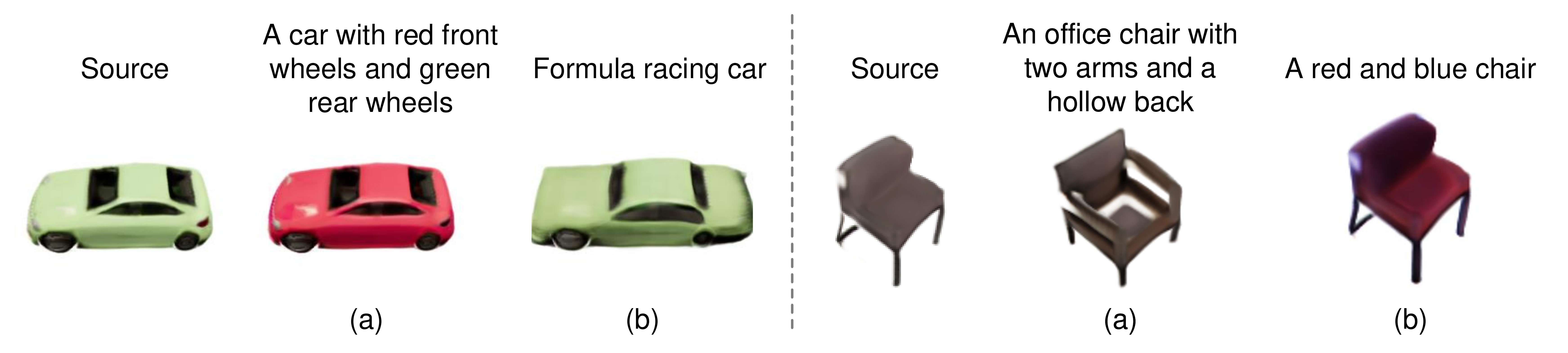}
  \caption{\textbf{Limitations.} Our method cannot handle fine-grained edits~(a) and out-of-domain edits~(b).}
  \label{fig:limitation}
\end{figure}

\begin{figure*}[t]
\centering
\includegraphics[width=1.0\linewidth]{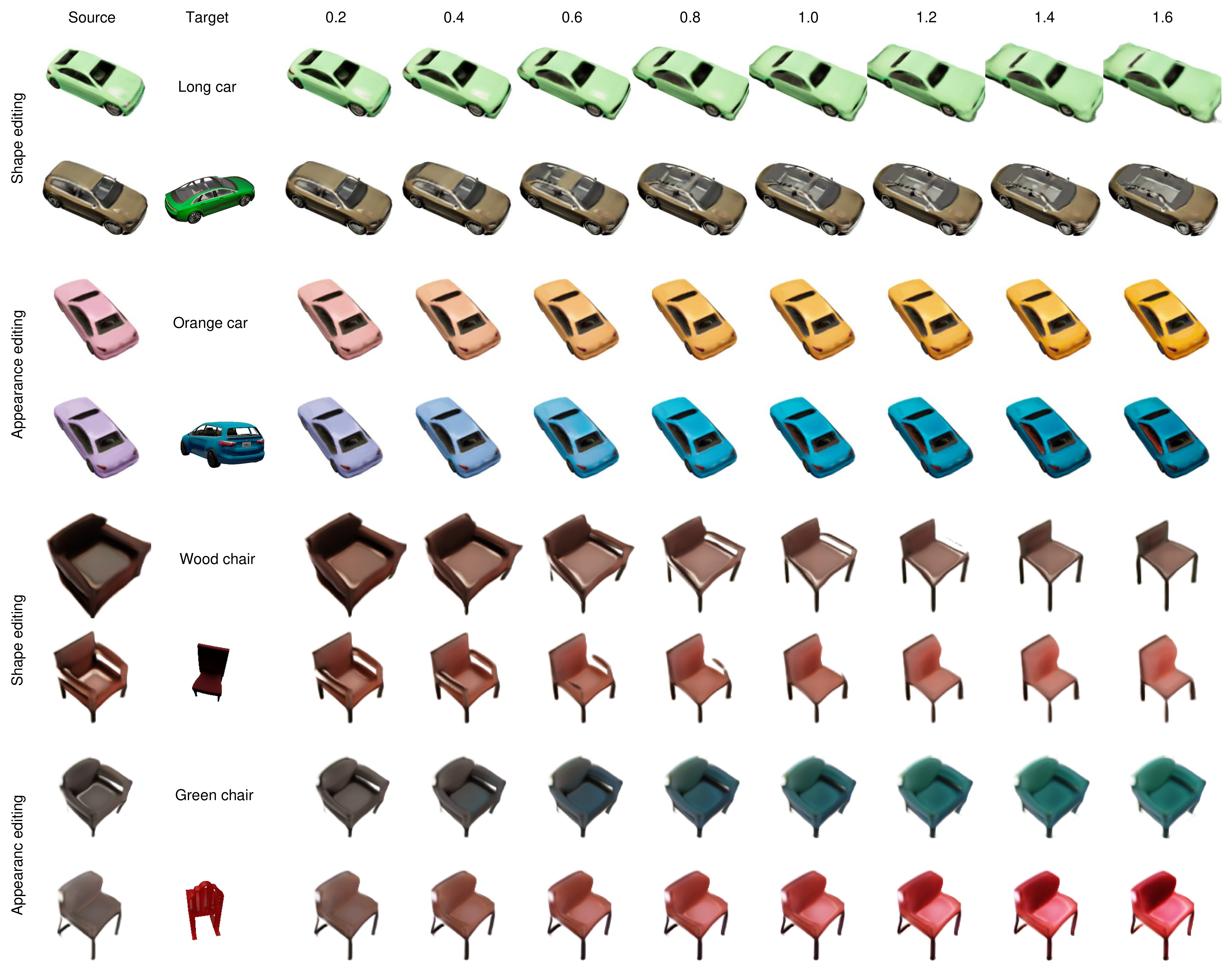}
   \caption{Editing results of moving along the $\Delta \vz_s$ and $\Delta \vz_a$ directions. The shape and appearance codes of the source are updated by adding $s *\Delta \vz_s$ and $s *\Delta \vz_a$, while $s$ is a scalar ranging from 0 to 1.6 with a step 0.2.}
   \label{fig:extend_delta}
\end{figure*}

\begin{figure*}[t]
\centering
\includegraphics[width=1.0\linewidth]{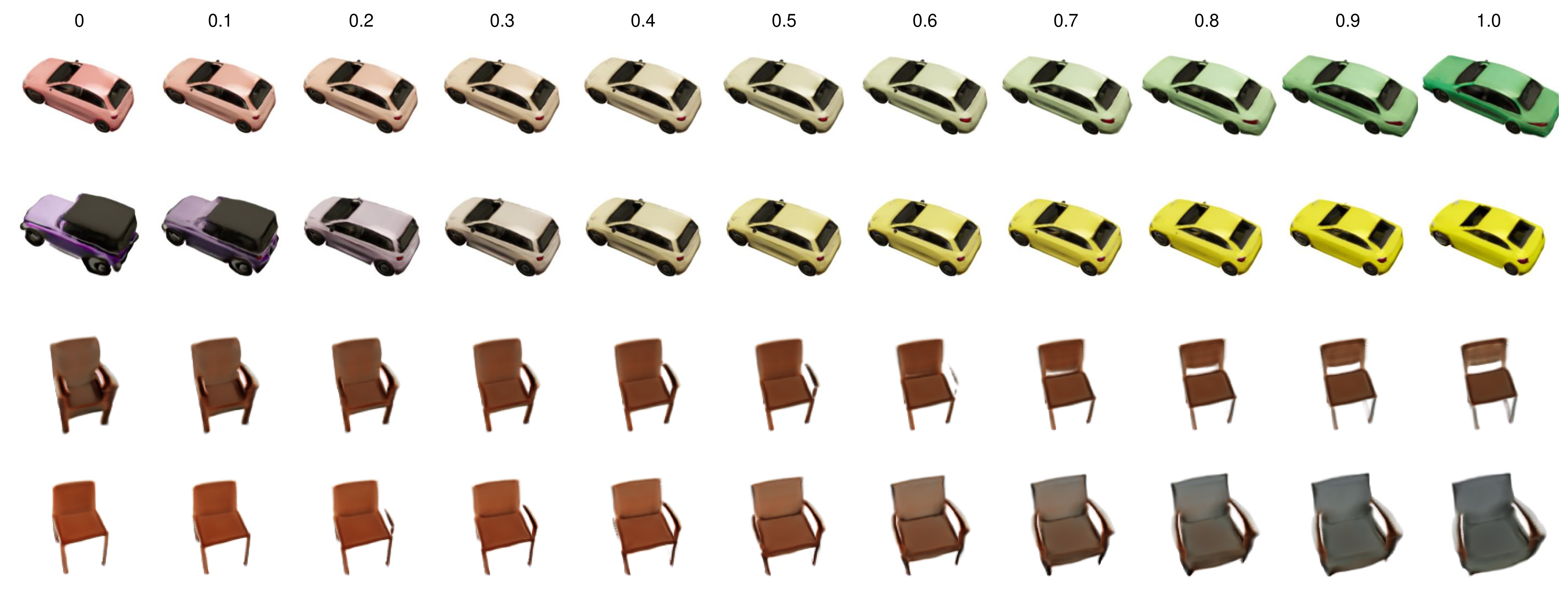}
   \caption{Interpolations in the latent space. The leftmost and rightmost ones are the inputs. And interpolation ratios are shown at the top.}
   \label{fig:extend_inter}
\end{figure*}

\begin{figure*}[t]
\centering
\includegraphics[width=1.0\linewidth]{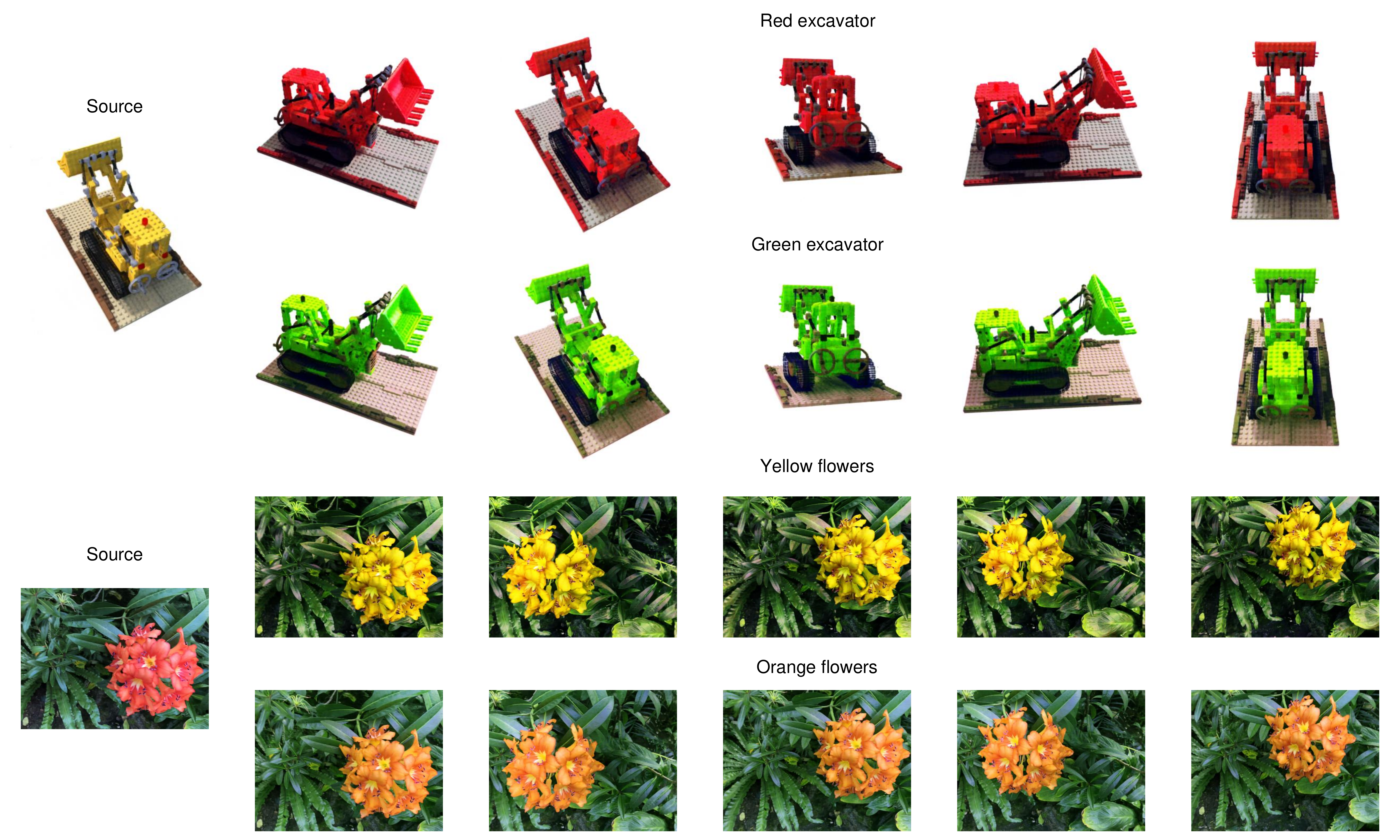}
   \caption{Single NeRF appearance editing results with our designed CLIP loss. NeRF models are trained on LLFF dataset~\cite{mildenhall2019local}.}
   \label{fig:extend_extra}
\end{figure*}

\section{Extended Discussions}

\textbf{Continuous Manipulation.} Our method supports editing both shape and appearance, given a single text prompt or an examplar. This can be achieved by continuously editing the shape and appearance, i.e., first editing the shape and then editing the color and vice versa. We show results in \Fref{fig:extend_contine}. This provides a user-friendly way for editing when users want to edit both the shape and appearance indicated by a single text description or an exemplar.

\textbf{Fine-grained appearance manipulation within a same color category}.
Though our method cannot handle fine-grained local parts shape and appearance edits as stated in the limitation, it supports fine-grained appearance manipulation at a whole object level, 
as shown in \Fref{fig:extend_fine}. Our method enjoys achieving various editing results within a same color category. Without loss of generality, we show various editing results related to the color blue.

\textbf{Scaling along Editing Direction.} From \eqref{eq:clip-mapper}, our code mappers provide manipulation directions $\Delta z_s=\big(\hat{\mathcal{E}}_t(\vt)\big)$ and $\Delta z_a=\appearmapper\big(\hat{\mathcal{E}}_t(\vt)\big)$ in the latent space for shape and appearance editing. 
We can scale along the editing direction to obtain gradually editing results through the following equation:
\begin{equation}
\begin{split}
\shapecode=s\times \Delta z_s+\shapecode',\\
\appearcode=s\times \Delta z_a+\appearcode',
\end{split}
\label{eq:clip-mapper}
\end{equation}
where $s$ is the scalar. This scaled scheme also supports directions learned from examplars. 
We show scaled manipulation results in \Fref{fig:extend_delta}. The manipulation effect becomes stronger as the scalar $s$ increases.

\textbf{Interpolation.} As shown in \Fref{fig:extend_inter}, the shape and appearance latent space supports interpolation between two latent codes $\vz^1=(\vz_s^1,\vz_a^1)$ and $\vz^2=(\vz_s^2,\vz_a^2)$. Given
an interpolation ratio $\vr$, we define the interpolated latent code $\vz_{inter}$ as $\vz_{inter}=\vz^2\times \vr + \vz^1\times (1-\vr)$, while $\vr$ ranges from 0 to 1.0 with a step 0.1. Then we can obtain the interpolated result using $\vz_{inter}$.

\textbf{Necessity of latent space.} Our method performs shape and appearance edits on the latent space of a conditional NeRF model with our designed CLIP constraints. A question arises whether the latent space is necessary, i.e.,  is it possible to edit the shape and appearance of a single NeRF model~\cite{mildenhall2020nerf} directly rather than a conditional NeRF? We first evaluate our designed CLIP loss on appearance editing in \Fref{fig:extend_extra}. Given a pre-trained NeRF model on the LLFF dataset~\cite{mildenhall2019local}, we fix the density-related layers and finetune the color-related layers of NeRF with our CLIP loss. We also use the patch-based ray samplar while calculating our CLIP loss. Our CLIP constraint succeeds in editing the color of a single NeRF model without any ground truth. However, we fail to achieve satisfying results while editing the shape of a single NeRF with our deformation network conditioned by a text prompt. This may be because the CLIP loss is still not strong and compact enough to deform the shape without the latent space constraint. We think it is an interesting problem to explore in the future. 

\section{Supplementary Video}
We provide a supplementary video with a real-time demo and more visual results rendered in multiple views. We highly recommend watching our supplementary video to observe the user-friendliness and view-consistency that our method can achieve in both shape and color editing.


\section{Conclusion}
We present the first text-and-image driven manipulation method for NeRF by designing a unified framework to provide users with flexible control over 3D content using either a text prompt or an exemplar image.
We design a disentangled conditional NeRF architecture that allows disentangling shape and appearance while editing an object, and two feedforward code mappers enable fast inference for editing different objects. Further, we proposed an inversion method to infer the shape and appearance codes from a real image, allowing editing the existing data.

\textbf{Limitations.}
We evaluate our approach by extensive experiments on various text prompts and exemplar images and provide an intuitive editing interface for interactive editing. However, our method cannot handle fine-grained and out-of-domain shape and appearance edits as shown in \Fref{fig:limitation}, due to the limited expressive ability of the latent space and the pre-trained CLIP. This may be alleviated by adding more various training data.

{\small
\bibliographystyle{ieee_fullname}
\bibliography{egbib}
}

\end{document}